\documentclass[pmlr,twoside,tablecaptionbottom]{pgm-jmlr}
\usepackage{pgm-pmlr-dat}
\firstpageno{1}
\usepackage[utf8]{inputenc}
\inputencoding{utf8}

\usepackage{booktabs}
\usepackage{multirow}
\usepackage{times}
\usepackage{marginnote}
\usepackage{comment}
\usepackage[linecolor=blue!60!,backgroundcolor=blue!10!,textwidth=3.0cm]{todonotes}\reversemarginpar

\input{settings.tex}

\title[Learning Non-Parametric Directed Acyclic Graphical Models]{Kernel-Based Differentiable Learning of Non-Parametric Directed Acyclic Graphical Models\titletag{}}
 \author{
 \Name{Yurou Liang\nametag{}} \Email{yurou.liang@tum.de}\\
 \Name{Oleksandr Zadorozhnyi} \Email{oleksandr.zadorozhnyi@tum.de}\\
 \Name{Mathias Drton} \Email{mathias.drton@tum.de}\\
 \addr Technical University of Munich  and Munich Center for Machine Learning, Germany 
 }

\begin{document}
\maketitle

\begin{abstract}
Causal discovery amounts to learning a directed acyclic graph (DAG) that encodes a causal model. This model selection problem can be challenging due to its large combinatorial search space, particularly when dealing with non-parametric causal models. Recent research has sought to bypass the combinatorial search by reformulating causal discovery as a continuous optimization problem, employing constraints that ensure the acyclicity of the graph. In non-parametric settings, existing approaches typically rely on finite-dimensional approximations of the relationships between nodes, resulting in a score-based continuous optimization problem with a smooth acyclicity constraint. In this work, we develop an alternative approximation method by utilizing reproducing kernel Hilbert spaces (RKHS) and applying general sparsity-inducing regularization terms based on partial derivatives. Within this framework, we introduce an extended RKHS representer theorem. To enforce acyclicity, we advocate the log-determinant formulation of the acyclicity constraint and show its stability. Finally, we assess the performance of our proposed RKHS-DAGMA procedure through simulations and illustrative data analyses.
\end{abstract}

\begin{keywords}
Causal discovery; graphical model; kernel methods; RKHS; structural equation model.
\end{keywords}

\section{Introduction}
\label{sec:intro}

Structural equation models (SEMs) based on directed acyclic graphs (DAGs) have found wide-spread applications ranging from computational biology \citep{zhang2023active} to manufacturing \citep{pmlr-v236-gobler24a} and finance \citep{ji2018network}.
To represent the joint dependence structure, each variable is modeled as a function of a subset of the other variables and noise.  In this setting, DAG-based models assume the absence of causal feedback loops. Although this assumption can be restrictive, it is crucial for defining non-linear models. Indeed, it is often unclear whether cyclic systems of structural equations have a unique solution or if they admit a solution at all.

In many applications, the underlying DAG is unknown, and methods for causal discovery, which learn the DAG from data, offer useful insights. Numerous algorithms have been proposed for causal discovery; see, e.g., \cite{drton:maathuis:2017} or \cite{spirtes:zhang:handbook}.  A classical constraint-based  approach relies on testing conditional independences \citep{spirtes1991algorithm,tsamardinos2003algorithms,margaritis1999bayesian}. Another prominent approach is score-based algorithms \citep{heckerman1995learning,chickering2002optimal}. In this work, we focus on a score-based approach.  Specifically, we will take up a recent theme that aims to find a DAG minimizing a model selection score through a continuous optimization problem with a continuous acyclicity constraint applied to a weighted adjacency matrix $W$.  This approach was initiated in the NO\-TEARS algorithm \citep{zheng2018dags}, which assumes a linear SEM and uses an exponential acyclicity constraint transforming the combinatorial optimization problem into a continuous one that is solved using an augmented Lagrangian scheme. 
Several follow-up works have proposed alternative characterizations of acyclicity \citep{yu2019dag,nazaret2023stable,ng2020role}. One such work introduced the DAGMA algorithm \citep{bello2022dagma}, which is generally faster than NOTEARS. In line with related literature, we refer to this broad approach as \textit{differentiable causal discovery}.

\cite{zheng2020learning} extended this methodology to non-parametric settings. 
Specifically, the authors modeled each variable as a non-parametric function of the other variables and noise, proposing to approximate the non-parametric functions by multi-layer perceptions or via a (truncated) basis expansion in the original functional space---i.e., the space of functions whose derivatives are square-integrable over the domain (an expansion is then possible via the trigonometric basis of functions). They then minimize the corresponding residual loss subject to the exponential acyclicity constraint. In our work, we build on this approach under additional assumptions that the non-parametric functions are continuously differentiable, taking up a perspective of (Gaussian) RKHS. 

In the MLP framework of \cite{zheng2020learning}, the entry \(W_{kj}\)  of the weighted adjacency matrix is defined as the \(L_2\) norm of the \(k\)th column of the weight matrix in the first hidden layer of the \(j\)th MLP. When approximating by basis expansions, the general non-parametric model is assumed to be an additive model. The entries of the weighted adjacency matrix are defined as the \(L_2\) norm of the coefficients corresponding to the basis approximation. 

Both the MLP and the basis expansion methods in \cite{zheng2020learning} lead to finite-dimensional optimization problems in terms of neural network weights or basis coefficients. 
The current MLP approximation is sensitive to the number of hidden units: while increasing the size of the hidden layers increases the flexibility of MLP functions, larger networks require more samples for accurate estimation \citep{zheng2020learning}. Moreover, the MLP approximation relies on random initialization of weights, which introduces randomness in results \citep[see  Figure~2 in][]{waxman2024dagma}. Fine-tuning the architecture of a neural network is, thus, a non-trivial task.  On the other hand, the current basis expansion approximations are restricted through a focus on additive models.

\paragraph{Contributions.} In this work, we present a novel kernel-based methodology for differentiable causal discovery in non-parametric settings.
Our contributions can be summarized as follows:
\begin{itemize}
    \item We approximate each non-parametric function that represents the dependency structure between random variables using an RKHS with a differentiable kernel $k$. We establish a version of an RKHS Representer Theorem for an empirical acyclicity-constrained optimization problem, similar to that of \citealt{rosasco2013nonparametric} in the statistical learning context.  Given data $(x^i)_{i=1}^n$, this leads to optimizing functions that are combinations of evaluations of the kernel and its partial derivatives:
    \begin{equation}\label{eq: sparse representer th}
  \sum_{i = 1}^{n}\alpha_i k(x, x^i) + \sum_{i = 1}^{n} \sum_{a = 1}^{d}\beta_{ai} \frac{\partial k(x, s)}{\partial s^a}\bigg\vert_{s=x^i}. 
	\end{equation} 
    \item Let $f_j$ represent the dependency structure between $j$-th random variable and the other random variables. If $x \mapsto f_{j}\paren{x}$ is continuously differentiable for all $x$ in a connected and compact sample space $\cX$, then $\frac{\partial f_j}{ \partial x_{k}} = 0$ implies that $f_{j}$ does not depend on $x_k$. Thus, we define the weighted adjacency matrix directly via the partial derivatives of the functions \(f_j\). This approach is model-agnostic in the sense that the weights do not refer to approximations to $f_j$. 
    \item We base our optimization on the DAGMA method   \citep{bello2022dagma} and adopt the log-determinant acyclicity constraint \(h_{\text{ldet}}\), for which we demonstrate stable optimization behavior on the boundary of the domain.
    \item We explore the behavior of kernel-based differentiable causal discovery in simulation experiments as well as the collection of cause-effect datasets from \cite{JMLR:v17:14-518}.  The code for our experiments is available on the first author's GitHub site.\footnote{ \url{https://github.com/yurou-liang/RKHS-DAGMA}}
\end{itemize}
\paragraph{Outline.} Section~\ref{sec: Background} sets up notation, reviews the DAG learning problem, and presents basic facts about RKHS. Section~\ref{sec: Acyclicity constraint} summarizes existing acyclicity constraints and discusses the stability 
of the log-determinant acyclicity constraint, on which our work is based. Section~\ref{sec: Overall learning objective} introduces the sparsity regularizer and examines the extended RKHS representer theorem for the constrained optimization problem with acyclicity constraint. 
Section~\ref{sec: Optimization} outlines the resulting constrained optimization problem and introduces the  RKHS-DAGMA algorithm for learning sparse nonparametric graphs. In Section~\ref{sec: Experiments}, we compare the performance of RKHS-DAGMA with different versions of nonparametric NOTEARS  \citep{zheng2020learning} in numerical experiments.
Additional details on the experiments and proofs are given in the Appendix.

\section{Background}
\label{sec: Background}

\subsection{Structural Equation Model and Differentiable Causal Discovery}
Let $X = (X_1, \dots, X_d)$  be a random vector taking values in $\cX \subset \mbr^{d}$ defined on some probability space $\paren{\Omega, \cF,\mbp}$. We assume $\cX$ is a bounded connected non-empty open set.
Let $[d] := \{ 1,2,\dots,d\}$. Consider a directed acyclic graph $G=(V,E)$ with vertex set $V = [d]$ and edge set $E \subset V \times V$.  As usual, $\mathrm{pa}(i)= \{j \in V: (j,i)\in E \}$ is the set of \textit{parents} of a vertex $i \in V$.  For a subset $A \subset [d]$, let $X_{A} = (X_{i})_{i \in A}$.  When $A=\emptyset$, we set $X_{A} \equiv 0$.

In a graphical model, each variable $X_j$ exhibits a constrained stochastic relationship with the other coordinates of random vector $X:=(X_j)_{j=1}^d$ \citep{handbook}.  Presenting the model through structural equations and assuming additive noise, the model for DAG $G$ postulates that 
\begin{align}\label{eq:model_assmpt}
    X_{j} = f_{j}(X) + \epsilon_{j},  \quad j=1,\dots,d,
\end{align}
where each measurable function $f_{j} : \mbr^{d} \mapsto \mbr$ depends only on the subvector $X_{\mathrm{pa}(i)}$, and $\paren{\epsilon_j}_{j \in [d]}$ are mutually independent stochastic error terms.  In this model, the conditional expectations are $\mbe[X_j| X_{\mathrm{pa}(j)}] = f_j(X)$, where we note again that $f_j$: $\cX \rightarrow \mbr$  does not depend on $X_k$ if $k \notin \mathrm{pa}(j)$.

The DAG learning problem for the considered models may be formulated as follows. Let $\mbx\in \mbr^{n \times d}$ be a data matrix whose rows $(x^{i})_{i=1}^n$ represent
 \(n\) i.i.d.~observations. 
Let $x_j^i$ be the $j$-th coordinate of the $i$-th observation. We denote the loss function by $\ell : \cX^n \times \cY^n \mapsto \mbr_{+} $, where $\cY \subset \mbr$ is the image space for predictions. The typical loss function is the least squares loss $\ell(y,\hat{y}) = \norm{y-\hat{y}}_2^2$. The goal is to estimate $f = (f_1,\dots, f_d)$ by minimizing the score function
\begin{align}\label{op}
    L(f) := \frac{1}{2n} \sum_{j = 1}^{d} \ell(X_{j}, f_j(\mbx)) \text{ subject to the dependencies in $f$ corresponding to a DAG}, 
\end{align}
where, in slight abuse of notation, $f_{j}\paren{\mbx} = \paren{f_{j}\paren{x^{1}},\ldots,f_{j}\paren{x^{n}}}$.
We assume that each \(f_j\) is continuously differentiable and that $f_j$ and its derivative are both square-integrable. 
As in \citet{waxman2024dagma}, we define the weighted adjacency matrix \(W \in \mbr^{d\times d}\) with entries \(W(f)_{kj} := \norm{\frac{\partial f_j\paren{\cdot}}{\partial x_{k}} }_{L^2}\). This definition is model-agnostic in that it does not refer to any particular family of approximation models for the function $f_j$ (as was done, e.g., in the MLP setup of \citealp{zheng2020learning}). 

In the sequel, $\mbp_X$ denotes the joint distribution of the random vector $X$.  We write $C^{1}\paren{\cX}$ for the space of continuously differentiable functions over $\cX$ and $L_{2}\paren{\cX}$ for the space of (equivalence classes of) square-integrable functions on $\cX$, i.e.,  functions $f:\cX\to\mbr$ with $\int_{\cX} f^{2}\paren{x}\mbp_X(dx)<\infty$. We will mostly drop dependence on the domain and on the underlying distribution $\mbp_{X}$ and simply write $L_2 := L_2(\cX)$ or $L_2 := L_2(\mbr^d)$ when the domain is clear from the context.
Furthermore, we use $\norm{\cdot}_{\infty}$ to denote the essential supremum norm w.r.t.~Lebesque measure $\lambda$ on a (subset) $\cX \subset \mbr$. For  $g: \cX \mapsto \mbr$ in $C^{1}\paren{\cX}$, we write $\frac{\partial}{\partial x_{k}}g(x)$ for its partial derivative (as a map $x \mapsto \frac{\partial}{\partial x_{k}}g\paren{x}$) and $\frac{\partial}{\partial x_{k}}g\paren{x} \vert_{x= s} $ for the derivative's value at point $x=s$. Finally, we denote the $i-$th smallest eigenvalue of a real matrix $A \in \mbr^{d \times d}$ by $\lambda_{i}\paren{A}$; so,  $|\lambda_1(A)|\leq |\lambda_2(A)|\leq\dots\leq |\lambda_d(A)|$.  The spectral radius \( \rho(A) = |\lambda_d (A)|\) is the largest eigenvalue of $A$ in magnitude.

\subsection{Kernels and RKHS}
To estimate the structural equation model in
\eqref{eq:model_assmpt},
we use the toolbox of kernel-based algorithms \citep[Chap.~4]{steinwart2008support}.  For a given vector space $\cX$, let $k : \cX \times \cX \mapsto \mbr$ be a symmetric function such that for any $n \in \mbn$, $\{x^{i}\}_{i=1}^{n} \in \cX ^{n}$, and $c \in \mbr^{n}$, it holds that $c^{\top}Kc \geq 0$, where $K = \paren{k\paren{x^i,x^j}}_{i,j = 1}^{n} \in \mbr^{n \times n}$ is the \textit{kernel matrix}. Such a function $k\paren{\cdot,\cdot}$ is termed a \textit{kernel}. Its values can be represented as an inner product in a Hilbert space $H_{0}$, i.e., $k\paren{x^i,x^j} = \inner{\phi(x^i),\phi(x^{j})}_{H_{0}},$ where $\phi : \cX \mapsto H_0$ is the \textit{feature map}. To every kernel $k$, we can associate a reproducing kernel Hilbert space (RKHS) $H$, which is a space of functions $f: \cX \mapsto \mbr$ for which the \textit{reproducing property} holds: $f\paren{x} = \inner{f, k\paren{\cdot,x}}_{H}$ for all $x \in \cX$, $f \in H$.  
\smallskip

Let \(H\) be an RKHS, and let $k$ being its correspondent reproducing kernel \(k : \cX \times \cX \mapsto \mbr \). Let \(\cR_{E, \cD}(f_j) := \frac{1}{n}\sum_{i =1}^{n} E(x_j^i, f_j(x^i))\) be the empirical risk for convex loss-function \(E : \mbr \times \cY \mapsto \mbr_{+}\). Consider the empirical risk with function complexity regularizer based on an arbitrary non-decreasing function $J : \mbr \mapsto \mbr_{+}$:
\begin{equation}\label{fct: loss}
    \cR_{E, \cD}(f_j) +  J\paren[1]{\norm{f_j}_H}.
\end{equation}
Then \citep[see, e.g.][ Thm 1]{Schoelkopf:2000}, the minimizer of \eqref{fct: loss} can be written as
\[\hat{f}_j(\cdot) = \sum_{i =1}^{n}\alpha^j_ik(\cdot, x^i),\]
where \(\alpha^j \in \mbr^n\). Furthermore, the reproducing property of the RKHS directly implies that $\hat{f}_{j}\paren{x}= \sum_{i =1}^{n}\alpha^j_ik(x, x^i)$ for all $x \in \cX$.

\section{Acyclicity Constraints}
\label{sec: Acyclicity constraint}

The earliest work that transfers the discrete acyclicity constraint in Problem \eqref{op} to a continuous differentiable constraint is NOTEARS \citep{zheng2018dags}. It introduced the following acyclicity constraint based on the trace-exponential of the weighted adjacency matrix \(W \in \mbr^{d \times d}\):
\[h_{\text{exp}} (W) = \tr( \exp(W \circ W)) -d, \] 
where \(\circ\) is the Hadamard product. Since $W \circ W$ has nonnegative entries, we may consider $h_{\text{exp}}$ also as a function of nonnegative matrices \(A \in \mbr_{\ge 0}^{d \times d}\), $A=W\circ W$.
Using the fact that $x \mapsto \exp(x)$ can be represented by uniformly over $\mbr$ convergent power series and by the linearity of the trace one can generalize the acyclicity constraint to the so-called \textit{Power Series Trace Family} (PST) \citep[see also][Def. 1]{nazaret2023stable}. The members of the family can be expressed as follows: 
\[h(A) = \sum_{k = 1}^{\infty} a_k \tr[A^k], \]
where \((a_k)_{k \in \mbn^*} \in \mbr_{\ge 0}^{\mbn^*}\), and \(A \in \mbr_{\ge 0}^{d \times d}\). 
 Table~\ref{tab:pst_constr} lists some examples.

\begin{table}[t]
\begin{minipage}{0.5\textwidth}
\centering
\begin{tabular}{llr}
\toprule
Name & $\alpha_k$ & $h_a$ \\
\midrule
$h_{\text{expm}}$ & $\frac{1}{k!}$ & $\tr\exp(A) - d$ \\
$h_{\text{log}}$ & $\frac{1}{k}$ & $-\log\det(I_d - A)$ \\
$h_{\text{inv}}$ & $1$ & $\tr(I_d - A)^{-1}$ \\
$h_{\text{binom}}$ & $\left( \begin{array}{c} d \\ k \end{array} \right)$ & $\tr(I_d + A)^d - d$ \\
\bottomrule
\end{tabular}
\captionof{table}{PST acyclicity constraints}
\label{tab:pst_constr}
\end{minipage}%
\hfill
\begin{minipage}{0.5\textwidth}
\centering
\begin{tabular}{llr}
\toprule
Name & $h_a$ \\
\midrule
$h_{\text{spectral}}$ & $|\lambda_d(A)|$ \\
$h_{\text{ldet}}$ & $-\log\det(sI_d - A) + d \log s$ \\
\bottomrule
\end{tabular}
\captionof{table}{Spectral-based acyclicity constraints}
\label{tab:spec_constr}
\end{minipage}
\end{table}

Another class of constraints, the \textit{spectral-based acyclicity constraints}, was developed based on the fact that the spectral radius of a non-negative weighted adjacency matrix \(A\) is zero if and only if \(A\) corresponds to a DAG \citep{nazaret2023stable,bello2022dagma}. Table~\ref{tab:spec_constr} gives examples.

\citet{nazaret2023stable} introduced three criteria for constraints to ensure stable optimization behavior and showed that the PTS acyclicity constraints are not stable during the optimization process. \citet{bello2022dagma} studied the acyclicity constraint $h_{\text{ldet}}$, based on the log-determinant of the weight-matrix $A$. More precisely, for $s > 0$, consider space of matrices $\mbw^s := \{A \in \mbr^{d \times d}:s>\rho(A)\}$ and define $h_{\text{ldet}}^{s}: \mbw^s \rightarrow \mbr$ as 
\begin{align}
\label{eq:log_determinant}
    h_{\text{ldet}}^{s}(A):= -\log\det (sI_d-A) + d\log s.
\end{align}
\citet{bello2022dagma} show favorable performance of $h_{\text{ldet}}$ in comparison to the constraints $h_{\text{exp}}$ and $h_{\text{poly}}$.  
We will adopt the constraint $h_{\text{ldet}}$ in the sequel. 
We explain our choice from the optimization perspective  in the following theorem.

\begin{theorem}\label{th: stability hldet}
    The constraint function $h_{\text{ldet}}^{s}\paren{\cdot}$ from \eqref{eq:log_determinant} satisfies the following stability properties:\\
    For all $A \in \mbr^{d\times d}_{\geq 0}$ it holds:
	\begin{itemize}
		\item \textbf{V-stable} If \( h_{\text{ldet}}^{s}(A) \neq 0 \), then 
        for $\epsilon \rightarrow 0^{+}$, \(h_{\text{ldet}}^{s}(\epsilon A) \ge c\epsilon\) for some positive constant \(c\).
		\item \textbf{D-stable} \( h_{\text{ldet}}^{s}(A) \) and \( \nabla h_{\text{ldet}}^{s}(A) \) are well-defined where $\triangledown h_{\text{ldet}}^{s}(A) = (sI_d - A)^{-T}$, with $\triangledown h_{\text{ldet}}^{s}(A) = 0$ if and only if $A$ is a DAG. 
	\end{itemize} 
 Finally, $h_{\text{ldet}}^{s}$ is an acyclicity constraint in the sense that
\[h_{\text{ldet}}^{s}(A) \geq 0 \text{ , with }h_{\text{ldet}}^{s}(A) = 0 \text{ if and only if graph generated by matrix} A \text{ is a DAG.}\]
\end{theorem}

No PST constraint is V-stable for $d \ge 2$, i.e., there exists $A \in \mbr^{d \times d}$, $h\paren{\epsilon A} =\cO\paren{\epsilon^d}$ as $\epsilon \rightarrow 0$, $\epsilon > 0$ \citep[Theorem 2]{nazaret2023stable}.  In contrast,  V-stability ensures $h_{\text{ldet}}^{s}$ does not vanish rapidly to $0$, and D-stability ensures that $h_{\text{ldet}}^{s}$ and its gradient exists. Although $h_{\text{spectral}}$ is both V-stable and D-stable \citep{nazaret2023stable}, we observe that in practice, 
$h_{\text{spectral}}(W)$ and $h_{\text{ldet}}^{s}(W)$ operate on distinctly different scales, with $h_{\text{spectral}}(W)$ being substantially larger than $h_{\text{ldet}}^{s}(W)$. This considerable difference in magnitude raises challenges in assessing whether $h_{\text{spectral}}(W)$ can be regarded as sufficiently small to be considered zero. Thus, we choose $h_{\text{ldet}}^{s}(W)$.

\section{Overall Learning Objective}
\label{sec: Overall learning objective}

\subsection{Sparsity Regularizer}

In applications, we often expect the random variables $X_j$ to have only a few parent variables and, thus, invoke a sparsity regularizer.
Recall that we assume that the functions $f_j$: $\cX \rightarrow \mbr$ are in the class $C^{1}\paren{\cX}$, and the functions $f_j$ and their derivatives are both square-integrable. We consider the regularizer (compare also \citealt{rosasco2013nonparametric})
\begin{align}\label{L2 sparsity regularizer}
    \Omega_1(f_j) = \sum_{k = 1}^{d}\norm{\frac{\partial f_j(\cdot)}{\partial x_k}}_{L_2} =  \sum_{k = 1}^{d}\sqrt{\int_{\cX} \paren{\frac{\partial f_j(x)}{\partial x_k}}^2 \mbp_X (dx)}. 
\end{align}

To develop a data-based decision rule we need to consider an empirical counterpart for the $\norm{\cdot}_{L^{2}}$ norm of the derivative. Thus,  we form an empirical estimate based on the data $\mbx= \{x^{i}\}_{i=1}^{n}$, by setting
\[\norm{\frac{\partial f_j(\cdot)}{\partial x_k}}_{n} := \sqrt{\frac{1}{n}\sum_{i = 1}^{n} \paren{\frac{\partial f_j(x^i)}{\partial x_k}}^2}.\]
Then the empirical estimate of \eqref{L2 sparsity regularizer} is
\[\Omega_1^{\cD}(f_j) = \sum_{k = 1}^{d} \norm{\frac{\partial f_j(\cdot)}{\partial x_k}}_{n}.\] 
Similarly, the empirical estimate of the coefficient $W_{jk}$ of the weighted adjacency matrix is 
\[W^{\cD}_{kj} = \norm{\frac{\partial f_j(\cdot)}{\partial x_k}}_{n}.\]

\subsection{Constrained Empirical Optimization Problem Solved by Kernel Methods}

For each \(j \in [d]\), we further assume \(f_j\): $\cX \rightarrow \mbr, j \in [d]$ to be in a reproducing kernel Hilbert space (RKHS) \(H\) generated by a bounded continuously-differentiable kernel \(k\) on \(\cX\) and use the term \(\lambda \norm{f_j}_H^2\) to penalize function complexity. Then we aim to minimize the following loss function: 
\begin{equation}\label{sparse loss fct}
    \sum_{j=1}^{d} \frac{1}{2n}  \ell(\mathbb{X}^{j}, f_j(\mbx)) + \tau(2\Omega_1^{\cD}(f_j) + \lambda \norm{f_j}_H^2) \text{ s.t. } h_{\text{ldet}}^{s}(W^{\cD}) = 0,
\end{equation}
where \(\tau,\lambda \) are positive numbers and $s>0$ is some fixed number (typically set to $1$). Following \citet{rosasco2013nonparametric}, we show a version of the RKHS Representer Theorem for problem \eqref{sparse loss fct} with the log-determinant acyclicity constraint. The main point of the result below is to show that the solution of the log-determinant constrained empirical minimization problem \eqref{sparse loss fct} admits the form of a finite linear combination, for which we may then optimize the weights.

\begin{theorem}
\label{th: sparse representer th}
    Let \(\cX\) be a bounded connected non-empty open set in \(\mbr^d\), \(k(\cdot, \cdot)\) be a bounded countiniously differentiable kernel.
    Then the constrained minimizer of \eqref{sparse loss fct} can be written as 
	\begin{equation}\label{eq: sparse representer th_func_j}
		\hat{f_j}^{\tau}(x) = \sum_{i = 1}^{n}\alpha_i^{j} k(x, x^i) + \sum_{i = 1}^{n} \sum_{a = 1}^{d}\beta_{ai}^{j} \frac{\partial k(x, s)}{\partial s_a}\bigg\vert_{s=x^i}, \quad x \in \cX,
	\end{equation} 
	where \(\alpha^{j}, (\beta_{ai}^{j})_{i = 1}^n \in \mbr^n \) and \(a, j \in [d] \). Then,
	\begin{equation}\label{eq: norm sparse representer th}
			\norm{\hat{f_j}^{\tau}}_H^2 = \sum_{i,l =1}^{n} \alpha_i^{j} \alpha_l^{j} k(x^i, x^l) + 2\sum_{i,l =1}^{n} \sum_{a = 1}^{d} \alpha_i^{j} \beta_{al}^{j}\frac{\partial k(x^i, x^l)}{\partial x^l_a} + \sum_{i,l =1}^{n} \sum_{a,b = 1}^{d} \beta_{ai}^{j}\beta_{bl}^{j} \frac{\partial k(x^i, x^l)}{\partial x^i_a \partial x^l_b}. 
	\end{equation}
\end{theorem}

Via Theorem \ref{th: sparse representer th}, we may estimate every function \(f_j\) by a kernel estimator. As no random variable should cause itself, \(f_j\) should not depend on \(x_j\) as its input.  Thus, to estimate $f_j$ we replace \(k\) in \eqref{eq: sparse representer th_func_j} with a restricted kernel \(k^{-j}\) that depends only on the subvector $(x_k)_{k\not=j}$ and use the representation
\[\hat{f_j}(x) = \sum_{i = 1}^{n}\alpha_i^{j} k^{-j}(x, x^i) + \sum_{i = 1}^{n} \sum_{a = 1}^{d}\beta_{ai}^{j} \frac{\partial k^{-j}(x, s)}{\partial s_a}\bigg\vert_{s=x^i}.\]
Let \(\theta_j = \{\alpha^{j}, \beta_{ai}^{j}: a \in [d], i \in [n]\}\) denote the parameters for each \(f_j\), and let \(\theta = (\theta_1, \dots, \theta_d)\). Then the loss function is constructed as follows:
\begin{equation}
\label{eq: loss function}
     \sum_{j = 1}^{d}  \frac{1}{2n} \ell(\mathbb{X}^{j}, \hat{f_j}^{\theta}(\mbx)) + \tau[2\Omega_1^\cD(\hat{f_j}^{\theta}) + \lambda \norm[1]{\hat{f_j}^{\theta}}_H^2].
\end{equation}
To evaluate the acyclicity constraint on the dataset, using the Representer Theorem for the function $\hat{f}_{j}$, we consequently obtain for every $k,j \in [d]$: 
\begin{align}
  W^{\cD}_{kj}(\hat{f_j}^{\theta})& = W^{\cD}(\theta)_{kj} = \norm{\frac{\partial \hat{f_j}^{\theta}(x)}{\partial x_k}}_{n} \nonumber \\
  & =\bigg \{ \frac{1}{n} \sum_{i = 1}^{n} \left[ \sum_{l = 1}^{n} \alpha_l^j \frac{\partial k^{-j}(x^i, x^l)}{\partial x^i_k} + \sum_{l = 1}^{n} \sum_{a = 1}^{d} \beta_{al}^j \frac{\partial k^{-j}(x^i, x^l)}{\partial x^i_k \partial x^l_a}\right]^2 \bigg \}^\frac{1}{2}, \label{eq: W}
\end{align}
and consequently
\begin{equation}\label{eq: sparsity}
  \Omega_1^{\cD}(\hat{f_j}^{\theta}) = \sum_{k=1}^{d} \sqrt{\frac{1}{n}\sum_{i = 1}^{n} \paren{\frac{\partial \hat{f_j}^{\theta}(x^i)}{\partial x_k}}^2 } = \sum_{k = 1}^{d}W^{\cD}(\theta)_{kj}. 
\end{equation}

As a default, we base our algorithm on the Gaussian kernel $k_{\gamma}\paren[1]{x,x'} = \exp(-\tfrac{1}{\gamma^2}\|x-x'\|^2)$, based on the Euclidean distance between $x$ and $x'$. For $\gamma >0$, $j \in [d]$, we define the restricted \textit{Gaussian kernel} \(k^{-j}_{\gamma} : \mbr^{d-1} \mapsto \mbr \) as 
\begin{equation}\label{eq:Gaussian_kernel}
    k^{-j}_{{\gamma}} (x, x') := \exp\bigg( -\frac{1}{\gamma^{2}} \sum_{i\neq j}^{d} (x_i - x'_i)^2 \bigg).
\end{equation}
It corresponds to the Gaussian kernel $k_{\gamma}\paren[1]{x,x'}$ when fixing the $j$-th coordinate to a constant. 

 \begin{remark}
    We want the decision rule returned by our estimation algorithm to belong to the space of continuously differentiable functions, as this guarantees that a vanishing partial derivative implies that the function does not depend on the considered coordinate.  To this end, we claim that any function that belongs to Gaussian RKHS $H_{\gamma}\paren{\cX}$ also belongs to $C^{1}\paren{\cX}$. To see this, let \(H_{\gamma}\) be the RKHS of the real-valued Gaussian RBF kernel \(k_{\gamma}\) for \(\gamma > 0\). Since $k_{\gamma}\paren{\cdot,\cdot}$ is (infinitely many times) differentiable, Theorem 10.45 in \citet{Wendlandt:2004} implies that the associated RKHS $H_{\gamma}\paren{\cX}$ is a subset of the space of continuously differentiable functions. Therefore,  $H_{\gamma}\paren{\cX} \subset C^{1}\paren{\cX}$, which implies that $\norm{f}_{C^{1}\paren{\cX}}< \infty$ for every $f\in H_{\gamma}$.   
 \end{remark}
 
\section{Optimization}
\label{sec: Optimization}

Combining Equations \eqref{eq: loss function}, \eqref{eq: W}, \eqref{eq: sparsity} together with least squares loss and log-determinant acyclicity constraint, we obtain the following constrained empirical optimization problem:
\begin{align*}
    &\min\limits_{\theta} \sum_{j = 1}^{d} \bigg \{ \frac{1}{2n} \norm{\mathbb{X}^{j} - \hat{f_j}^{\theta}(\mbx)}_2^2 + \tau[2\Omega_1^{\cD}(\hat{f_j}^{\theta}) + \lambda \norm[1]{\hat{f_j}^{\theta}}_H^2] \bigg \}\\
   &\text{s.t. }-\log \det(sI_d - W^{\cD}(\theta) \circ W^{\cD}(\theta)) + d\log (s) = 0.
\end{align*}
As in DAGMA, we use a central path method to solve the constrained optimization problem. We give our method in \algorithmref{alg: RKHS-DAGMA} and refer to it as RKHS-DAGMA. Note that line \ref{subopt} of \algorithmref{alg: RKHS-DAGMA} means that starting at $\theta = \theta^{(t)}$, $\theta^{(t+1)}$ is obtained by the ADAM optimizer \citep{kingma2014adam}.
\begin{algorithm2e}
\caption{RKHS-DAGMA}
\label{alg: RKHS-DAGMA}
 \DontPrintSemicolon
 \LinesNumbered
 \KwIn{Data matrix \(\mbx\), initial coefficient (learning step) $\mu^{(0)}$ (e.g., $1$), decay factor $\alpha \in (0,1)$(e.g., $0.1$), sparsity parameter $\tau$(e.g., $1\times 10^{-4}$), function complexity parameter $\lambda$ (e.g., $1\times 10^{-3}$), log-det parameter $s > 0$ (e.g., $1$), number of iterations $T$ (e.g., $6$), threshold $\omega$ (e.g., $0.1$).}
 \KwOut{$\hat{W}$, the estimated weighted adjacency matrix.}
 \BlankLine
Initialize $\theta^{(0)}$ so that $W^{\cD}(\theta^{(0)}) \in \mbw^s.$  \\
\For{$t \leftarrow 0$ \KwTo $T-1$}{
  Starting at $\theta^{(t)}$, solve $\theta^{(t+1)} = \argmin_{\theta}\mu^{(t)}(\sum_{j = 1}^{d} \{\frac{1}{2n} \norm[1]{\mathbb{X}^{j} - \hat{f_j}^{\theta}(\mbx)}_2^2 + \tau[2\Omega_1^{\cD}(\hat{f_j}^{\theta}) + \lambda \norm[1]{\hat{f_j}^{\theta}}_H^2]\}) + h_{\text{ldet}}^s(W^{\cD}(\theta)).$\label{subopt}\\
  Set $\mu^{(t+1)} = \alpha \mu^{(t)}.$
}

Threshold matrix  $\hat{W} = W^{\cD}(\theta^{(T)})\cdot \1(W^{\cD}(\theta^{(T)}) > \omega)$.
\end{algorithm2e}

\section{Experiments}
\label{sec: Experiments}

This section is divided into three parts. First, we analyze the performance of RKHS-DAGMA in a simple bivariate prediction setting,  distinguishing cause-effect within artificially constructed toy models. Second, we evaluate and compare the properties of RKHS-DAGMA with non-parametric NOTEARS algorithms, including NOTEARS-MLP and NOTEARS-SOB, as well as the score-based FGES algorithm \citep{DBLP:journals/corr/Ramsey15a} on sampled directed Erd\H{o}s-R\'enyi graphs of growing dimension. Finally, we assess the performance of RKHS-DAGMA against NOTEARS-MLP, NOTEARS-SOB, and FGES on real-world bivariate datasets \citep{JMLR:v17:14-518}. 

\subsection{Toy Example}
\begin{figure}[t]
\floatconts
  {fig: toy example}
  {\caption{Illustrations of RKHS-DAGMA by toy examples with two nodes.}}
  {%
    \subfigure[quadratic relationship]{\label{fig:quadratic}%
      \includegraphics[width=0.48\linewidth]{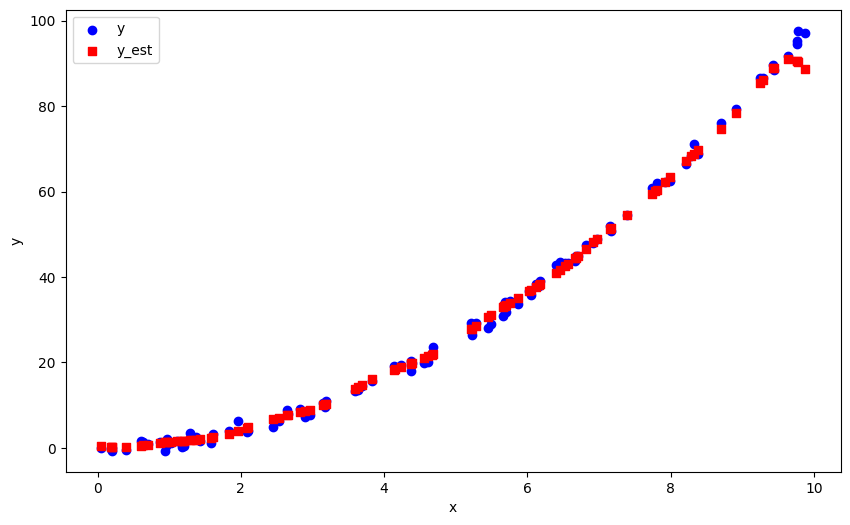}
      }%
    \;
    \subfigure[sine relationship]{\label{fig:sin}%
      \includegraphics[width=0.48\linewidth]{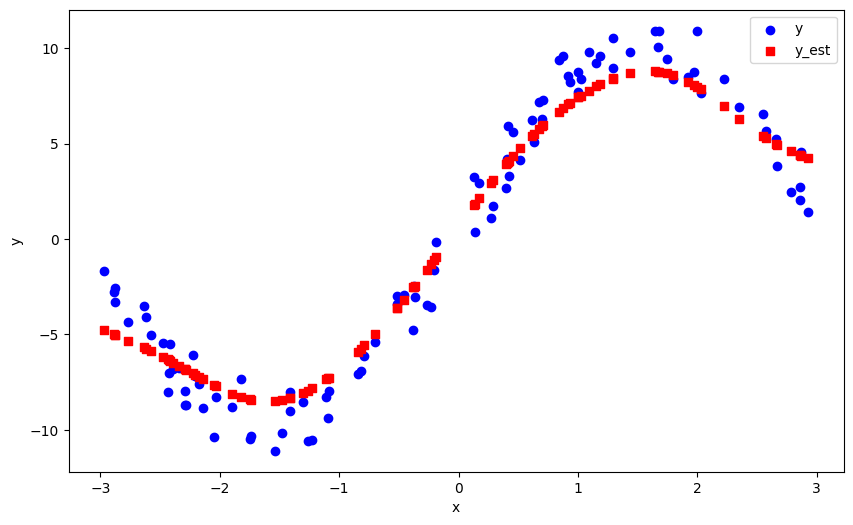}}
  }
\end{figure}

We illustrate the performance of RKHS-DAGMA by two simple simulations
with $d=2$ nodes and $n=100$ data points.  \figureref{fig: toy
  example} plots the ground truth data points (blue) and estimated
function values obtained  by RKHS-DAGMA (red) in the bivariate causal
models (a) $Y = X^{2} + \epsilon$, $X \sim \cU[0,10]$ and (b) $Y = 10\sin(X) + \epsilon$, $X \sim \cU[-3,3]$.  Let \(W_{\text{est}}\) be the estimated weighted adjacency matrix without any thresholding. 
  The results of \figureref{fig:quadratic} and \figureref{fig:sin} correspond to the estimated matrices
  \[W_{\text{est}} = 
  \begin{pmatrix}
        0 & 10.35\\
        6.22 \times 10^{-4} & 0
    \end{pmatrix},
    \qquad
    W_{\text{est}} = \begin{pmatrix}
        0 & 4.91\\
        8.49 \times 10^{-4} & 0
    \end{pmatrix},\]
    respectively.
    In both cases, \( W_{12}\) is large, whereas \(W_{21}\) is small enough to be ignored after thresholding, indicating that RKHS-DAGMA finds correct causal relationships.
    
\subsection{Structure Learning}

Next, we compare RKHS-DAGMA to  non-parametric NOTEARS methods by
comparing the estimated DAG with the ground truth, generated as
Erd\H{o}s-R\'enyi (ER) directed graphs with given topological ordering.
\citet{zheng2020learning} consider several graph models, including
ER and scale-free graphs. As one of the hardest settings \citep[Fig.~4]{zheng2020learning}, we take up ER$4$ graphs
for which NOTEARS algorithms are less competitive compared to
algorithms like fast greedy equivalence search
\citep{ramsey2017million}, DAG-GNN \citep{yu2019dag}, or greedy
equivalence search with generalized scores
\citep{huang2018generalized}.  ER$m$ ($m=4$) denotes
an ER graph with $m\times d$ edges.

Given the ground truth, we simulate from an SEM with \(X_j = g_j(X_{\mathrm{pa}(j)}) + \epsilon_j\) with \(\epsilon_j \sim \cN(0,1)\). We consider the following functional relationships for \(g_j\): Additive models with Gaussian processes, Gaussian processes, and MLPs according to the procedure of \citet{zheng2020learning}.  We also add a simulation type called the combinatorial model, which is an additive model with non-linear relationships random picked from the following common non-linear functions: \(g(x) = \exp(-|x|), \;g(x) = 0.05 \cdot x^2, \;g(x) = \sin(x)\). We refer to Appendix~\ref{apd:Experiments Details} for comprehensive details on the simulations.

As noted in \citet{bello2022dagma}, the initial point \(W(\theta^{(0)})\) is required to be inside \(\mbw^s\). Since the zero matrix is always inside \(\mbw^s\) for any \(s > 0\), we set the parameters \(\theta^{(0)}\) be 0. Given that our approximation method and sparsity regularizer fundamentally differ from those in NOTEARS algorithms, the hyperparameters $\lambda$ and $\tau$ are tuned via grid search. In RKHS-DAGMA, we take sparsity parameter \(\tau = 1\times10^{-4}\), function complexity parameter \(\lambda = 1\times10^{-3}\), and threshold \(\omega = 0.1\). Additionally, we take \(\mu^{(0)} = 1\) and the default value \(T = 6\); if the resulting weighted adjacency matrix is not a DAG, we enhance \(T\) to \(7\). We set \(\gamma = 0.4d\) for the Gaussian kernel (see Appendix~\ref{rm: gamma} for an explanation of this choice). Due to the explicit computation of derivatives and the Hessian of the kernel function, we limit the maximum number of iterations of the ADAM optimizer to \(10\%\) of corresponding values in DAGMA to compensate for the additional cost. For the NOTEARS algorithms, we choose the default hyperparameters as described in \citet{waxman2024dagma} and \citet{zheng2020learning}.

To evaluate model performance, we use the structural Hamming distance
(SHD), which measures the total number of edge additions, deletions,
and reversals needed to convert the estimated graph into the true
graph. A lower SHD indicates better model performance.

Our results indicate that RKHS-DAGMA consistently outperforms
NOTEARS-SOB in terms of structured Hamming distance
(SHD). Additionally, compared to NOTEARS-MLP, RKHS-DAGMA shows
superior performance in simulations based on Gaussian processes (GP),
additive GP, and combinatorial models, while maintaining competitive
results in MLP experiments (see \figureref{fig: model comparison1}-\ref{fig: model comparison2}).

\begin{figure}[t]
\vspace{-15pt}
\floatconts
  {fig: model comparison1}
  {\caption{Comparison between RKHS-DAGMA and NOTEARS-MLP by SHD (lower is better) for random data generated from \ref{fig:gp1} the ER-$4$ GP model, \ref{fig:gp-add1} the ER-$4$ GP-additive model, \ref{fig:mlp1} the ER-$4$ MLP model, \ref{fig:combi1} the model with combination of functions. Boxplots show the median and quartiles across $10$ different simulations for each simulation model.}}
  {%
    \subfigure[\text{}]{\label{fig:gp1}%
      \includegraphics[width=0.45\linewidth]{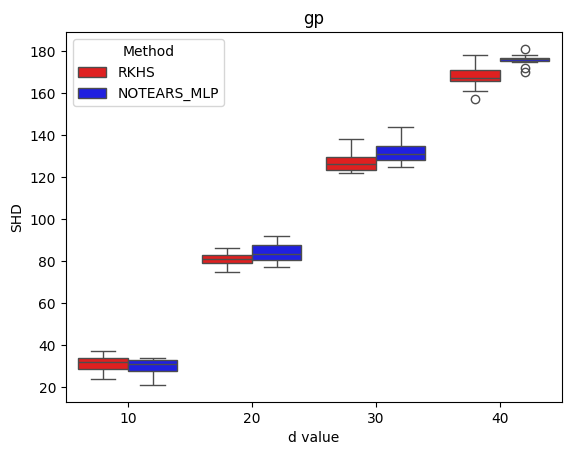}
      }%
    \;
    \subfigure[\text{}]{\label{fig:gp-add1}%
      \includegraphics[width=0.45\linewidth]{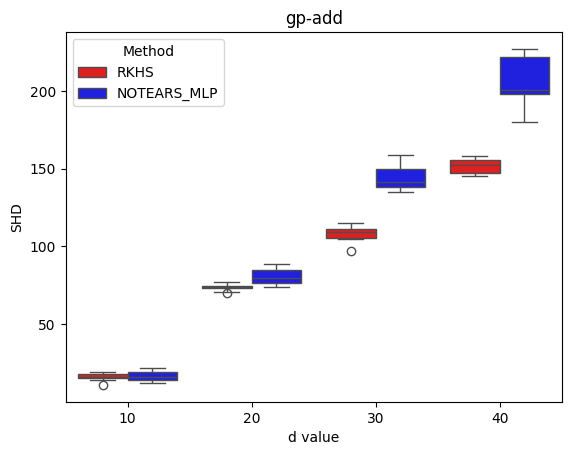}
  }\\
  
     \subfigure[\text{}]{\label{fig:mlp1}%
      \includegraphics[width=0.45\linewidth]{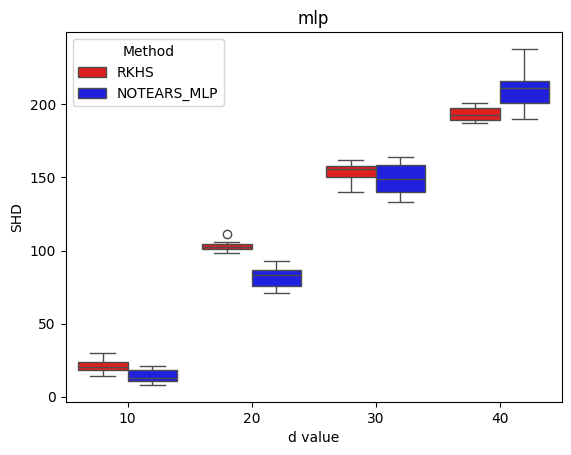}
      }
      \;
     \subfigure[\text{}]{\label{fig:combi1}%
      \includegraphics[width=0.45\linewidth]{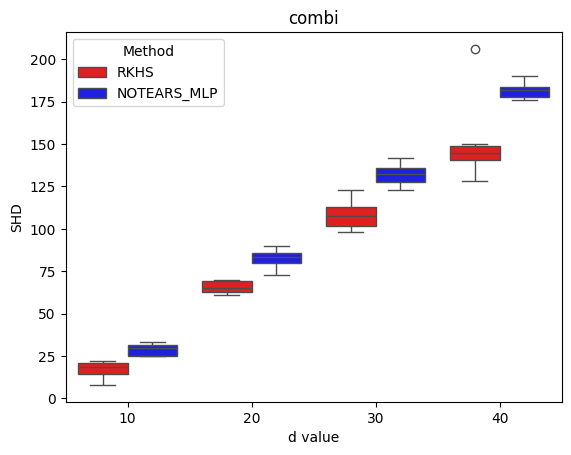}
      }
  }
\end{figure}
\begin{figure}[t]
\vspace{-15pt}
\floatconts
  {fig: model comparison2}
  {\caption{Comparison between RKHS-DAGMA and NOTEARS-SOB by SHD (lower is better) for random data generated from \ref{fig:gp2} the ER-$4$ GP model, \ref{fig:gp-add2} the ER-$4$ GP-additive model, \ref{fig:mlp2} the ER-$4$ MLP model, \ref{fig:combi2} the model with combination of functions. Boxplots show the median and quartiles across $10$ different simulations for each simulation model.}}
  {%
     \subfigure[\text{}]{\label{fig:gp2}%
     \includegraphics[width=0.45\linewidth]{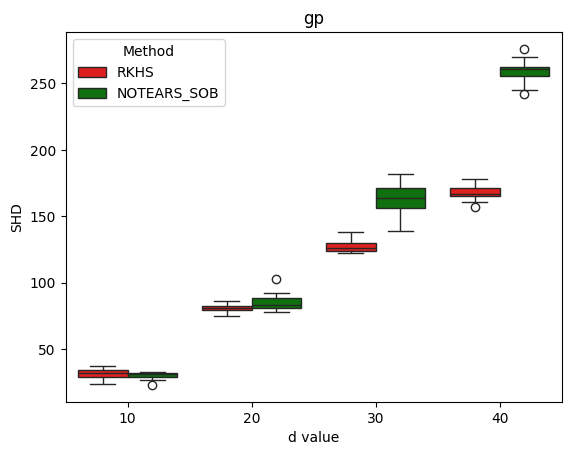}
      }%
      \;
    \subfigure[\text{}]{\label{fig:gp-add2}%
      \includegraphics[width=0.45\linewidth]{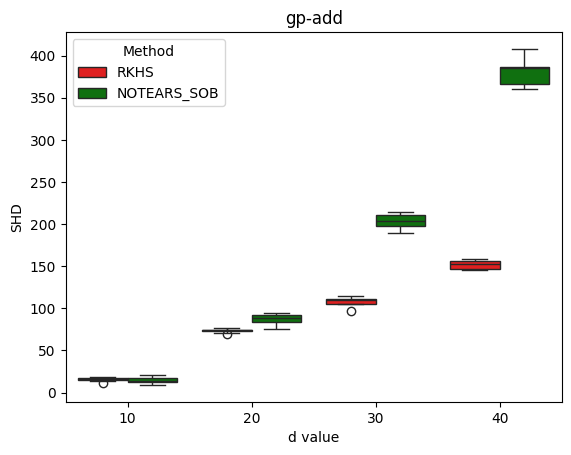}
      }\\
     \subfigure[\text{}]{\label{fig:mlp2}%
      \includegraphics[width=0.45\linewidth]{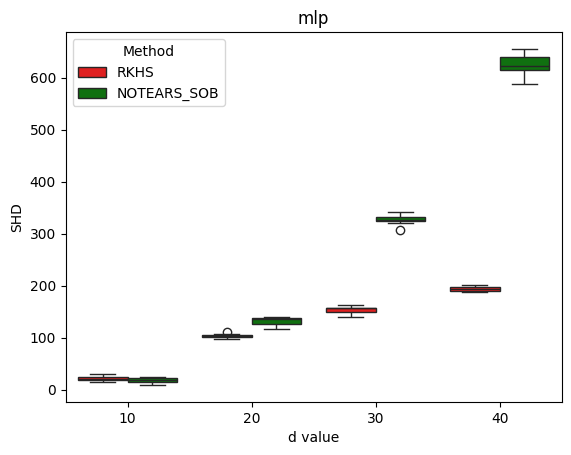}
      }
      \;
     \subfigure[\text{}]{\label{fig:combi2}%
      \includegraphics[width=0.45\linewidth]{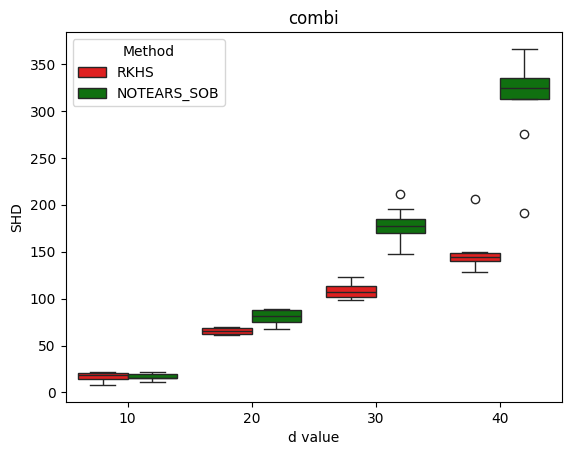}
      }
  }
\end{figure}

\begin{figure}[t]
\vspace{-15pt}
\floatconts
  {fig: model comparison3}
  {\caption{Comparison between RKHS-DAGMA and FGES by SHD (lower is better) for random data generated from \ref{fig:gp3} the ER-$4$ GP model, \ref{fig:gp-add3} the ER-$4$ GP-additive model, \ref{fig:mlp3} the ER-$4$ MLP model, \ref{fig:combi3} the model with combination of functions. Boxplots show the median and quartiles across $10$ different simulations for each simulation model.}}
  {%
     \subfigure[\text{}]{\label{fig:gp3}%
     \includegraphics[width=0.45\linewidth]{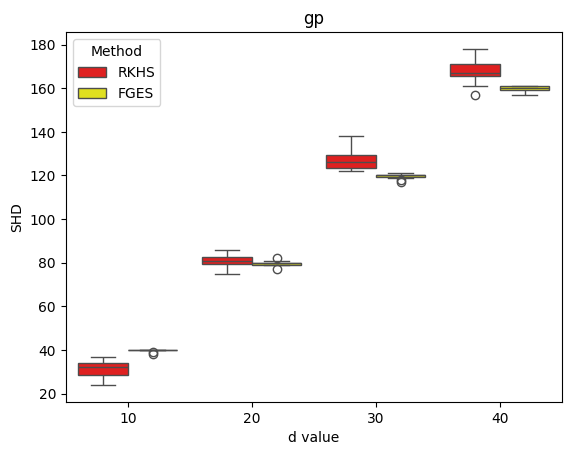}
      }%
      \;
    \subfigure[\text{}]{\label{fig:gp-add3}%
      \includegraphics[width=0.45\linewidth]{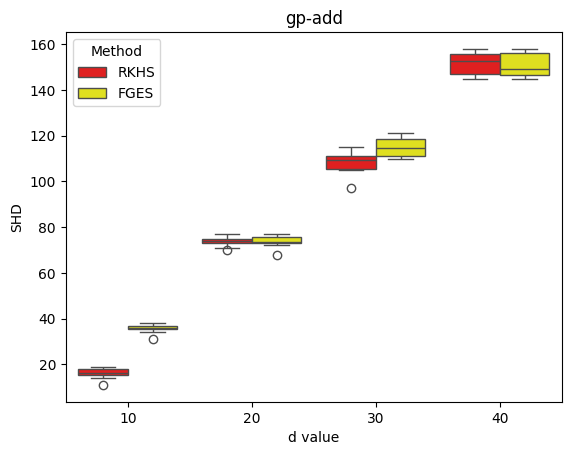}
      }\\
     \subfigure[\text{}]{\label{fig:mlp3}%
      \includegraphics[width=0.45\linewidth]{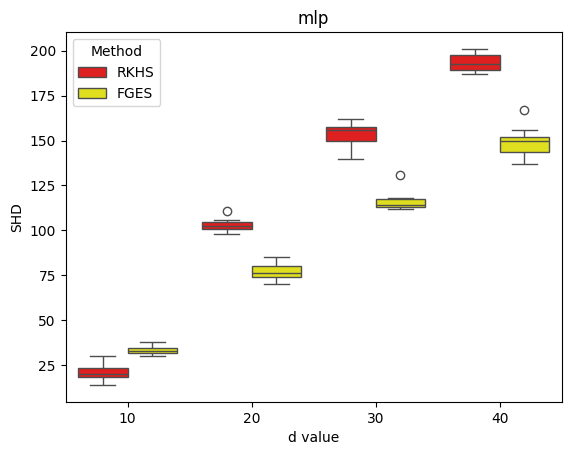}
      }
      \;
     \subfigure[\text{}]{\label{fig:combi3}%
      \includegraphics[width=0.45\linewidth]{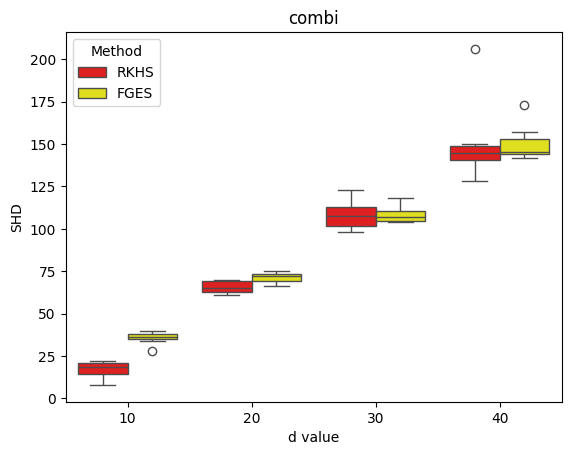}
      }
  }
\end{figure}

We also conduct a comparative analysis between RKHS-DAGMA and the FGES algorithm, implemented through the \texttt{py-causal}
package \footnote{\url{https://github.com/bd2kccd/py-causal}}.
\figureref{fig: model comparison3} indicates that RKHS-DAGMA
significantly outperforms the FGES algorithm in simulations based on GP,
additive GP, and combinatorial models for  $d = 10$. It 
achieves comparable performance $d \ge 20$. However, in the MLP
simulations, the performance of RKHS-DAGMA clearly falls behind for $d\ge 20$.

\subsection{Real Data}
\label{subsec:cause_effect_data}

Finally, we compare the performance of RKHS-DAGMA with NOTEARS-MLP, NOTEARS-SOB, and the FGES on a benchmark collection of datasets featuring cause-effect pairs \citep{JMLR:v17:14-518}.
These datasets are bivariate, each consisting of one pair of statistically dependent variables.  We excluded six datasets containing multi-dimensional random variables and standardized the remaining datasets.
 To reduce computational costs, datasets with more than 400 samples were organized in ascending order according to the first variable. These datasets were then partitioned into 300 grids according to the first variable, with the median value of the first variable in each grid and the corresponding value of the second variable used for model evaluation.
 
  RKHS-DAGMA achieves the best accuracy of $55.88\%$ among the remaining $102$  datasets, while NOTEARS-SOB and NOTEARS-MLP achieve an accuracy of $45.10\%$ and $0.98\%$ correspondingly. We attribute the bad performance of NOTEARS-MLP to the small sample size with a relatively large number of hidden units compared to the number of nodes, and to the specific definition of the weighted adjacency matrix, which depends on the weights of the first hidden layer and may differ significantly from those defined by derivatives \citep{waxman2024dagma}. Furthermore, the FGES algorithm yields only undirected edges across all bivariate datasets, as it employs a conditional independence test to eliminate unnecessary edges. In the context of bivariate data, where each variable pair lacks additional conditioning variables, the algorithm's capacity to extract information is inherently limited.
  
\section{Conclusion}
In this work, we addressed the non-parametric DAG learning problem using a procedure that exploits the machinery of infinite-dimensional (Gaussian) RKHS. We showed in Theorem \ref{th: sparse representer th} that the RKHS-DAGMA Algorithm, which solves a (combined) constrained empirical optimization problem with log-determinant acyclicity constraint, admits an explicit solution as a finite-dimensional representation of the kernel elements of the data and their derivatives. Furthermore, this solution can be computed using central path methods similar to those in the \textsc{DAGMA} algorithm. We compared the accuracy of RKHS-DAGMA with the known baselines under non-parametric structural equation modeling, such as \textsc{NOTEARS-MLP} and \textsc{NOTEARS-SOB} (see ex. \citealp{zheng2020learning}), as well as the score-based FGES algorithm \citep{DBLP:journals/corr/Ramsey15a} under settings comparable to those in \citet{zheng2020learning}. The RKHS-DAGMA Algorithm demonstrates utility across both simulated and empirical datasets.

\acks{Yurou Liang is supported by the DAAD programme Konrad Zuse Schools of Excellence in Artificial Intelligence, sponsored by the Federal Ministry of Education and Research.}

\newpage
\bibliography{references}
\newpage
\appendix

\section{Detailed Proofs}\label{apd:first}

\subsection{Proof of Theorem \ref{th: stability hldet}}
    \begin{itemize}
    \item \textbf{D-stable:} Follows from Theorem 1 in \citet{bello2022dagma}.
    \item \textbf{V-stable:} 
    Note that 
    \[\log\det(sI-A)-d \log s = \log(s^d\det(I-s^{-1}A))-d \log s = \log\det(I-s^{-1}A).
    \]
    Since \(s > \rho(A)\) is equivalent to \(1>\rho(s^{-1}A)\), we consider the case \(s=1\) w.l.o.g.
    For  any $\epsilon>0$ with $\epsilon \abs{\rho\paren{A}} \leq 1 $, it holds that
    \begin{align*}
        h_{\text{ldet}}^s(\epsilon A) &= -\log\det(I-\epsilon A) = -\log( \prod_{i = 1}^{d} \lambda_i(I-\epsilon A)) = \sum_{i = 1}^{d}-\log(1-\epsilon\lambda_i(A)).
    \end{align*}
We prove the statement of the Theorem under the assumption that the eigenvalues $\{\lambda_{i}\paren{A}\}_{i \in [d]}$ of $A$ are complex numbers and under the additional assumption that $\sum_{i=1}^{d}\lambda_{i}\paren{A} \neq 0$.

 Applying the Cauchy-integral formula to the principal branch of the complex logarithm function $z \mapsto \log\paren{1-z}$, which is analytic within domain $\abs{z} < 1$, we get: 
\[
  -\log\paren{1 - \epsilon \lambda_{i}\paren{A} } = \epsilon \lambda_{i}\paren{A} +\frac{\epsilon^{2}\lambda^{2}_{i}\paren{A}}{2\pi i}\int_{\eta_{0}}\frac{\log\paren{1-w}}{\paren{\epsilon\lambda_{i}\paren{A}-w} w^{2}} \,dw,  
\]
where $\eta_{0}$ is any closed circle of radius $r_{0}$ with $\epsilon \abs{\lambda_{i}\paren{A}} < r_{0} < 1$. Therefore, summing over all complex values $\lambda_{i}\paren{A}$ and using triangle inequality $\abs{a-b} \geq \abs{a} - \abs{b}$ for $a,b \in \mbc$, we deduce that
\[
  \abs{ \sum_{i=1 }^{d} -\log\paren{1 - \epsilon \lambda_{i}\paren{A} }} \geq  \epsilon \abs{\sum_{i=1}^{d} \lambda_{i}\paren{A}} - \frac{1}{2 \pi}\abs{ \sum_{i=1}^{d}{\epsilon^{2}\lambda^{2}_{i}\paren{A}}\int_{\eta_{0}}\frac{\log\paren{1-w}}{\paren{w-\epsilon\lambda_{i}\paren{A}} w^{2}} \,dw}. 
\]
Since $w \mapsto \frac{\log\paren{1-w}}{w - \epsilon \lambda_{i}\paren{A}}$ is continuous, Weierstrass Theorem implies that the function is bounded on the bounded domain $\eta_{0}$, yielding that $\abs{\frac{\log\paren{1-w}}{w-\epsilon \lambda_{i}\paren{A}}} \leq K$ for some $K>0$. Since $\frac{\epsilon \abs{\lambda_{i}\paren{A}}}{r_{0}} < 1$ and $\eta_{0}$ is a circle of radius $r_{0}$, the ML inequality for the complex integral gives
\[
\abs{{\epsilon^{2} \lambda^{2}_{i}\paren{A}}\int_{\eta_{0}}\frac{\log\paren{1-w}}{w - \epsilon \lambda_{i}\paren{A} w^2} \,dw} \leq 2 \pi \epsilon^{2} \lambda^{2}_{i}\paren{A} \frac{K}{r_0},
\]
which in turn implies that
\begin{align*}
\abs{ \sum_{i=1 }^{d} -\log\paren{1 - \epsilon \lambda_{i}\paren{A} }} &\geq  \epsilon \abs{\sum_{i=1}^{d} \lambda_{i}\paren{A}} - \frac{K}{r_{0}}\epsilon^2\abs{\sum_{i=1}^{d} \lambda^{2}_{i}\paren{A} }\\ 
                        & = \epsilon \abs{\sum_{i=1}^{d} \lambda_{i}\paren{A}} - \frac{K}{r_{0}}\epsilon^{2}\sum_{i=1}^{d}\lambda^{2}_{i}\paren{A} \\
                        & = \epsilon \abs{\sum_{i=1}^{d}\lambda_{i}\paren{A}} - \frac{K\epsilon^2 \norm{A}^{2}_{2}}{r_{0}} \\
                        & \geq \epsilon \frac{\abs{\sum_{i=1}^{d}\lambda_{i}\paren{A}}}{2},
\end{align*}
where the last inequality holds provided $\epsilon$ is chosen small enough, with 
\[\epsilon \leq \min\bigg \{\frac{1}{\abs{\rho\paren{A}}},\frac{r_{0}\abs{\sum_{i=1}^{d}\lambda_{i}\paren{A}}}{2K \norm{A}_{2}^{2}}\bigg\}.
\]
Thus, we proved the claim. 

\end{itemize}
Finally, notice that \citet{bello2022dagma} showed that $h_{\text{ldet}}^s$ is an acyclicity constraint and the instability of PST acyclicity constraint is shown in \citet{nazaret2023stable}.

\subsection{Proof of Theorem \ref{th: sparse representer th}}
\begin{proof}
	\eqref{eq: sparse representer th_func_j}: 
 Consider arbitrary elements $f,g \in H$. Since  $H$ is complete, there exists a sequence  $\paren{g}_{n=1}^\infty$, $g_{n} \in H$,  that converges to $g$ in the norm of Hilbert space $H$. Furthermore, for any $n \in \mbn$,
	\begin{equation*}
		 \inner{f,g_{n}}_{H} - \inner{f,g}_{H} \leq |\langle f, g_n \rangle_H - \langle f, g \rangle_H| = |\langle f, g_n-g \rangle_H| \overset{\text{Cauchy-Schwarz}}{\leq} \norm{f}_H\cdot\norm{g_n - g}_H.
	\end{equation*}
    Since the kernel \(k\) is bounded, elements $f\in H$ are bounded \citep[Lemma~4.23]{steinwart2008support}. Furthermore, since \(g_n \rightarrow g\) in the norm of the space $H$ we have that 
    \[
        \lim_{n \rightarrow \infty} \inner{f,g_{n}}_{H} - \inner{f,g} = 0, 
    \]
which implies that map $x \mapsto \inner{\cdot,x}_{H}$ is continuous.  Similarly, one can prove that the scalar product is continuous in the first coordinate.

For coherence of the remainder of the proof, we first refine the proof that for the open set $\cX \subset \mbr^{d}$, we have that $\frac{\partial}{\partial x_{a}} k\paren{\cdot, x} \in H $ for every $a \in [d]$ and $x \in \cX$,  and that  a ``differential reproducing property'' holds, which states that 
 \[\frac{\partial }{\partial x_{a} } f\paren{x} = \inner{f , \frac{\partial }{\partial s_{a}} k\paren{\cdot,s}\bigg \vert_{s = x} }_{H}, \quad\forall x \in \cX, \quad  f \in H.\]
 Specifically, we refine the proof of Theorem 1 a)-b) in
 \cite{zhou:2008} in case $\alpha =0$ to show that the ``derivative
 element'' exists in $H$. We deviate from that proof in the part that
 establishes the differential reproducing property, the difference
 being the use of completeness of $H$ and the fact that weak
 convergence and pointwise convergence are equivalent in RKHS $H$.  

Consider arbitrary $x \in \cX$. As $\cX$ is open, there exists $r > 0$
such that the ball
$\{x+y,y\in \mbr^d, \norm{y}_{2} \leq r\} \subset \cX$. For $a \in
[d]$, let $e_a$ be  the $a$-th orthonormal vector in the standard
Euclidean basis in $\mbr^{d}$.  Let  $\tilde{h}_{x,a} : \cX \mapsto
\mbr$  be the function  given by $\tilde{h}_{x,a}\paren{y} =
\frac{\partial}{\partial s_{a}}k\paren{s,y}\vert_{s = x}$ for all $y
\in \cX$.
By definition of the RKHS, $k\paren{\cdot, x} \in H$, and the set
of functions $H \mapsto \mbr$ given by
\begin{align}
\label{eq:weakly_comp_set}
\bigg \{ \frac{1}{t} \paren{k\paren{\cdot,x+te_{a}} - k\paren{\cdot,x}}: \abs{t} \leq r \bigg \}
\end{align} 
satisfies that for every $t, \abs{t} \leq r$: 
\begin{align*}
\norm{\frac{k\paren{\cdot, x + t e_{a}} - k\paren{\cdot, x}}{t}}_{H}^{2} &= \frac{1}{t^{2}} \paren[2]{k\paren{x+te_{a},x+te_{a}} - k\paren{x,x+t e_{a}} - k\paren{x+t e_a,x} + k\paren{x,x} } \\ 
&  \leq \norm{ \frac{\partial^{2}}{\partial x_{a} \partial x_{a}} k\paren{\cdot,\cdot}}^{2}_{\infty},
\end{align*}
where the last inequality follows from the fact that
$k\paren{\cdot,\cdot}$ is a continuously differentiable function in
every coordinate and an application of the Mean Value Theorem (twice,
once in every coordinate). The latter inequality implies that set
\eqref{eq:weakly_comp_set} lies within a ball of radius $\norm{
  \frac{\partial^{2}}{\partial x_{a} \partial x_{a}}
  k\paren{\cdot,\cdot}}_{\infty}$ in Hilbert space $H$. It is known
(since $\star$-weak convergence is equivalent to weak convergence in
Hilbert spaces) that the ball in the Hilbert space is weakly
sequentially compact; see, e.g., \cite[Chap.~3]{Rudin:1991}. Thus, there exists a sequence $\paren{t_{n}}$ such that $\lim_{n \rightarrow \infty }t_{n} = 0$ and the sequence $\frac{1}{t_{n}}\paren{k\paren{\cdot,x+t_{n}e_a} - k\paren{\cdot,x}}$ converges weakly to an element $h_{x} \in H$. The latter means that 
\begin{align}
        \label{eq:weak_limit}
        \lim_{n \rightarrow \infty}
  \inner{\frac{1}{t_{n}}\paren{k\paren{\cdot,x+t_{n}e_a} -
  k\paren{\cdot,x}}, f}_{H} = \inner{h_{x},f}_{H} \qquad\forall f \in H. 
\end{align}

Consider $f(\cdot)= k\paren{\cdot, y}$, where $y \in \cX$ is
arbitrary. Since $k\paren{\cdot,y}$ is differentiable (as a function
of the first coordinate), the left-hand side of~\eqref{eq:weak_limit} satisfies
\begin{align*}
\lim_{n \rightarrow \infty} \inner{\frac{1}{t_{n}}\paren{k\paren{\cdot,x+t_{n}e_a} - k\paren{\cdot,x}}, k\paren{\cdot,y}}_{H} &= \lim_{n \rightarrow \infty} \frac{1}{t_{n}}\paren{k\paren{x+t_{n}e_{a},y} - k\paren{x,y}}\\
&= \frac{\partial}{\partial s_{a}} k\paren{s,y}\bigg \vert_{s = x} = \tilde{h}_{x,a}\paren{y}.
\end{align*}
But for the right-hand side of~\eqref{eq:weak_limit}, it holds that 
$$\lim_{n \rightarrow \infty}
\inner{\frac{1}{t_{n}}\paren{k\paren{\cdot,x+t_{n}e_a} -
    k\paren{\cdot,x}}, k\paren{\cdot,y}}_{H} =
\inner{h_{x},k\paren{y,\cdot}} = h_{x}\paren{y}\qquad \forall y \in
\cX.$$
We conclude that $\tilde{h}_{x,a} = h_{x}$ as a map $\cX \mapsto
\mbr$, and since $h_{x}$ is in $H$, so is $\tilde{h}_{x}$. By
identifying $\frac{\partial}{\partial x_{a}} k\paren{\cdot,x} :=
h_{x}$, the existence of an element $y$ with $\frac{\partial}{\partial x_a} k\paren{x,y} = \inner{\frac{\partial}{\partial x_a} k\paren{x,\cdot},k\paren{\cdot, y}}$ follows. 

Now we show the differentiable reproducing property, that is, the convergence to the limit $\frac{\partial}{\partial x_{a}}\paren{k\paren{\cdot,x}}$ is pointwise (and not only as a weak limit) and we can exchange the differential and inner product sign. The latter is equivalent to the folklore fact that weak convergence is equivalent to pointwise convergence when the underlying space is RKHS. Indeed, consider any sequence $g_{n}$ that converges weakly to an element $g \in H$, which means that
$\lim_{n \rightarrow \infty} \inner{g_{n},f} = \inner{g,f}$ for all
$f \in H$. In particular, the latter holds for all $k\paren{x,\cdot}$,
$x \in \cX$ yielding the necessity.  To show the sufficiency, if
$\lim_{n \rightarrow \infty } g_{n}\paren{x} = g(x)$ for all $x \in X$
then $\lim_{n \rightarrow \infty} \inner{g_{n},f} = \inner{g,f}$ for
all $f \in \text{span}\{k\paren{x,\cdot}\}$, by linearity of the inner
product and its continuity. The claim then follows since $H$ is
complete. From this statement, we deduce that the limit in
\eqref{eq:weak_limit} is actually a pointwise limit.  Thus, for every
$x \in \cX$, we have
$\lim_{t \rightarrow 0}\frac{1}{t}\paren{k\paren{\cdot,x+te_a} -
  k\paren{\cdot,x}} = \frac{\partial}{\partial x_{a}} k
\paren{\cdot,x}$ and, moreover, it holds for every $f \in H$ that
\[
\inner{\frac{\partial}{\partial x_{a}} k \paren{\cdot,x},f}_{H} = \inner{\lim_{t \rightarrow 0}\frac{1}{t}\paren{k\paren{\cdot,x+te_a} - k\paren{\cdot,x}},f} = \lim_{t \rightarrow 0 } \frac{f\paren{x+te_a}-f\paren{x}}{t} =  \frac{\partial f\paren{x}}{\partial x_{a}},
\]
where we used continuity of inner product and the reproducing property
in the second equality.  Hence, the derivative exists, and the differential reproducing property holds. 

Let \(X' := \{x^i:i = 1, \dots, n\}\), and let \[
  H|_{X'} :=
  \text{span} \left\{k(\cdot, x^i), \frac{\partial k(\cdot,
      s)}{\partial s_a}|_{s =x^i}: i = 1,\ldots, n, \;a = 1, \ldots,
    d\right\}.\]
Furthermore, let \(H_{|X'}^{\perp}\) be the orthogonal complement of
\(H|_{X'}\) in \(H\); it exists and is well-defined as
every element in the span exists and is well-defined.
By the Hilbert Projection Theorem, every $f_{j} \in H$, $j \in [d]$ can be uniquely decomposed as 
\(f_{j} = f_{j}^{\parallel} + f_{j}^{\perp}\), where $f_{j}^{\parallel} \in H|_{X'}$ and $f_{j}^{\perp} \in H_{|X'}^{\perp}$.
	Let $e_a \in \mbr^d$ be the $a$-th vector of the standard Euclidean
 basis in $\mbr^d$.  By the reproducing property and the definition of
 $H_{|X'}^{\perp}$ in $H$, it holds that
 \[
 f_{j}^{\perp}(x^i) =  \inner {f_{j}^{\perp}, k(\cdot, x^i)}_{H} = 0.
 \]
Using the fact (proved above) that the differentiation reproducing
property holds for $f_{j} \in H$, together with the orthogonal property we get: 
\[
\frac{\partial f_{j}^{\perp}}{\partial x_{a}}\paren{x^{i}} = \inner{f_{j}^{\perp}, \frac{\partial k(\cdot, s)}{\partial s_a}|_{s = x^i}}_{H}  \overset{f_{j}^{\perp} \in H_{|X'}^{\perp}}{=} 0. 
\]

By the reproducing property in $H$, we deduce that for every $j \in [d]$, it holds 
\[
\frac{1}{n}\ell\paren{\mathbb{X}^{j}, {f_j}\paren{\mbx}}^{2} =  \frac{1}{n}\ell\paren[2]{\mathbb{X}^{j}, {f_j}^{\parallel}\paren{\mbx}}^{2}, 
\]
whereas the differential reproducing property implies that it holds that
\begin{align*}
W^{\cD}_{aj}(f) &= \frac{1}{n} \sum_{i = 1}^{n} \frac{\partial f^{2}_j(x^i)}{\partial x_a^i} = \frac{1}{n} \sum_{i = 1}^{n} \frac{\paren{\partial f^{\parallel}_j(x^i)}^{2}}{\partial x_a^i} = W^{\cD}_{aj}(f^{\parallel})
\end{align*}
Notice further that 
\[
\norm{\frac{\partial f_{j}(x)}{\partial x_a}}_{n}  = \norm{\frac{\partial f_{j}^{\parallel}(x)}{\partial x_a}}_{n},
\]
which in turn implies that $\Omega_1^{\cD}(f_{j}) = \Omega_1^{\cD}(f_{j}^{\parallel})$. 
	Thus,  by denoting 
 \[
 \cR_{L,\cD} (f) + \tau(2\Omega_1^{\cD}(f) + \lambda\norm{f}_{H^{d}}^2) = 
 \sum_{j = 1}^{d} \bigg \{ \frac{1}{2n} \norm{\mathbb{X}^{j} - {f_j}(\mbx)}^2 + \tau[2\Omega_1^{\cD}({f_j}) + \lambda \norm[1]{{f_j}}_H^2] \bigg \},
 \]
 for $f = \paren{f_{1},\ldots,f_{d}} \in H^{ \otimes d}$, where $H^{ \otimes d}$ denotes the direct product of $d$ copies of $H$, we get that over the acyclicity constraint  the following chain of the equalities holds:
	\begin{align*}
		&\inf\limits_{f \in H^d, h^s_{\text{ldet}}(W^{\cD}(f)) = 0} \cR_{L,\cD} (f) + \tau(2\Omega_1^{\cD}(f) + \lambda\norm{f}_{H^{d}}^2)\\
		&=\inf\limits_{\substack{f \in H^d, f = f^{\parallel} + f^{\perp}, \\f^{\parallel}\in H|_{X'}^d, h^s_{\text{ldet}}(W^{\cD}(f^{\parallel})) = 0}} \cR_{L,\cD} (f^{\parallel}) + \tau(2\Omega_1^{\cD}(f^{\parallel}) + \lambda\norm[1]{f^{\parallel}}_{H^{d}}^2 + \lambda\norm[1]{f^{\perp}}_{H^{d}}^2)\\
		&= \inf\limits_{\substack{f \in H^d, f = f^{\parallel} + f^{\perp}, f^{\parallel}\in H|_{X'}^d,\\ \norm{f^{\perp}}_{H^{d}} = 0, h^s_{\text{ldet}}(W^{\cD}(f^{\parallel})) = 0}} \cR_{L,\cD} (f^{\parallel}) + \tau(2\Omega_1^{\cD}(f^{\parallel}) + \lambda\norm[1]{f^{\parallel}}_{H^{d}}^2)\\
		&= \inf\limits_{f \in H|_{X'}^d, h^s_{\text{ldet}}(W^{\cD}(f)) = 0} \cR_{L,\cD} (f) + \tau(2\Omega_1^{\cD}(f) + \lambda\norm{f}_{H^{d}}^2).
    \end{align*}
    Thus, we showed that
   \[ 
   \inf\limits_{\substack{f \in H^d, \\ h^s_{\text{ldet}}(W^{\cD}(f)) = 0}} \cR_{L,\cD} (f) + \tau(2\Omega_1^{\cD}(f) + \lambda\norm{f}_{H^{d}}^2) = \inf\limits_{\substack{f \in H|_{X'}^d, \\ h^s_{\text{ldet}}(W^{\cD}(f)) = 0}} \cR_{L,\cD} (f) + \tau(2\Omega_1^{\cD}(f) + \lambda\norm{f}_{H^{d}}^2),
   \]
   establishing the first claim of the Theorem holds.
   \medskip
   
	To prove \eqref{eq: norm sparse representer th}, we first note that, by the differential reproducing property,
	\begin{equation}
		\langle \frac{\partial k(\cdot, x)}{\partial x_a}, \frac{\partial k(\cdot, y)}{\partial y_b} \rangle_{H} = \frac{\partial^2 k(x,y)}{\partial x_a \partial y_b} \label{eq: scalar product partial k}.
	\end{equation}
  Plugging in the formula for the solution of the constrained minimization problem and using reproducing and differential reproducing properties we obtain: 
	\begin{align}
 \nonumber
			\norm{ \hat{f_j}^{\tau}}_H^2 & = \inner{\sum_{i=1}^n \alpha_i^j k\left(\cdot, x^i\right)+\sum_{i=1}^n \sum_{a=1}^d \beta^j_{a i} \frac{\partial k(\cdot, x^i)}{\partial x_a^i}, \sum_{i=1}^n \alpha^j_i k\left(\cdot, x^i\right)+\sum_{i=1}^n \sum_{a=1}^d \beta^j_{a i} \frac{\partial k(\cdot, x^i)}{\partial x_a^i}}_H
   \\ 
\nonumber
			& =\sum_{i,l=1}^n \alpha^j_i \alpha^j_l k\left(x^i, x^l\right)+2\sum_{i, l=1}^n \sum_{a=1}^d \alpha^j_i \beta^j_{a_l}\inner{\frac{\partial k\left(\cdot, x^l\right)}{\partial x_a^l}, k\left(\cdot, x^i\right)}_H \\
\nonumber
			& +\sum_{i, l=1}^n \sum_{a, b=1}^d \beta^j_{a i} \beta^j_{b l}\inner{\frac{\partial k\left(\cdot, x^i\right)}{\partial x_a^i}, \frac{\partial k\left(\cdot, x^l\right)}{\partial x_b^l}}_H \\
			& =\sum_{i, l=1}^n \alpha^j_i \alpha^j_l k\left(x^i, x^l\right)+2\sum_{i, l=1}^n \sum_{a=1}^d \alpha_i^j \beta_{a l}^j \frac{\partial k\left(x^i, x^l\right)}{\partial x_a^l}+\sum_{i, l=1}^n \sum_{a, b=1}^d \beta_{a i}^j \beta_{b l}^j \frac{\partial k\left(x^i, x^l\right)}{\partial x_a^i \partial x_b^l}, 
		\end{align}
 which finalizes the proof. 
\end{proof}

\section{Details of Numerical Experiments and Intuition for the Choice of the Bandwidth of the Gaussian Kernel $\gamma$. }\label{apd:Experiments Details} 

\subsection{Simulations.}

We simulate ER4 DAGs following the procedure outlined by
\citet{zheng2020learning} with the functional relationship established in
three distinct ways and add an additional simulation type.

\begin{enumerate}
  \item
The first way employs the Gaussian process: 
\[f_j(X) = g_{j}(X_{\mathrm{pa}(j)}) + \epsilon_j \quad\forall\; j \in [d],\]
where \(g_{j}\) is sampled from RBF GP with lengthscale $1$. In
detail, for the sub-data matrix $\mbx_{\mathrm{pa}(j)} \in
\mbr^{n\times p}$ where $p \leq d$, we have $g_j(X_{\mathrm{pa}(j)}) \sim \cN(0, K(\mbx_{\mathrm{pa}(j)}, \mbx_{\mathrm{pa}(j)}))$ with 
\[K(\mbx_{\mathrm{pa}(j)}, \mbx_{\mathrm{pa}(j)})_{a,b}= \exp{\paren{-\frac{\norm{\vec{x}^a-\vec{x}^b}_2^2}{2l^2} }}.\] 
Here, $\vec{x}^a, \vec{x}^b$ denote the $a$-th and $b$-th row of
$\mbx_{\mathrm{pa}(j)}$, respectively, and the lengthscale is $l =
1$. Moreover, \(\epsilon_j \sim \cN(0,I_n)\) is a standard Gaussian
noise.

\item The second way is referred to as the additive Gaussian process (additive GP). It sets
\[f_j(X) = \sum_{k \in \mathrm{pa}(j)}^{} g_{kj}(X_k) + \epsilon_j,\]
where each \(g_{kj}\) is sampled from RBF GP with lengthscale $1$.
\item The third way involves using an MLP network with a hidden layer size of 100
  and a sigmoid activation function, where all weights are sampled
  from \(\cU((-2.0, -0.5) \cup (0.5, 2.0))\).
\item Moreover, we introduce an
  additional simulation type called the combinatorial model, where the
  non-linear relationship is a linear combination of various common
  non-linear functions: 
\[f_j(X) = \sum_{k \in \mathrm{pa}(j)}^{} g_{kj}(X_k) + \epsilon_j,\]
where \(g_{kj}\) is randomly picked from following non-linear functions: 
\[g(x) = \exp(-|x|), \;g(x) = 0.05 x^2, \;g(x) = \sin(x).\]

\end{enumerate}

\subsection{Hyperparameter choice}\label{rm: gamma}

While considering the experimental setting with different dimensions
of the underlying Erd\H{o}s-R\'enyi graphs, we notice that the
the complexity of the decision rule increases with the grows of the dimension. Thus, one observes a ``classical'' phenomenon of the curse of
dimensionality \citep[see for example][] {giraud:2020,
  gyorfi:2002}. In a nutshell, high-dimensional i.i.d.~observations
are ``essentially'' equidistant from each other, while the distance
between the points grows with the growing dimension, which poses a
problem for high-dimensional metric-based methods. To handle this
problem, we seek to reduce the
estimation error for the signal by decision rules with higher
regularity.

In our case, we employ the machinery of RKHS rules based on the
Gaussian kernel \eqref{eq:Gaussian_kernel} with parameter $\gamma$.
Let $H_{\gamma}$ be the Gaussian RKHS
with reproducing kernel
$k\paren{x, \cdot} := k_{\gamma}\paren{x, \cdot} =
\exp\paren{-\frac{\norm{x-\cdot}^{2}}{\gamma^2}} $. It then holds that $H_{\gamma_2} \subset H_{\gamma_1}$ for
$\gamma_2 > \gamma_1 >0 $ 
\citep[see][Proposition~4.46]{steinwart2008support}.  Hence, higher
regularity results from a larger choice of $\gamma$. Notice that for bounded domains
$\cX$, the same effect (i.e., restricting to the spaces of larger
smoothness) can be ensured by considering the re-scaled domain with
parameter $\frac{1}{\gamma}$.
\end{document}